\pdfoutput=1
\documentclass[11pt]{article}

\usepackage{emnlp2021}

\usepackage{times}
\usepackage{latexsym}
\usepackage{multirow}
\usepackage{graphicx}
\usepackage{epigraph} 
\usepackage{algorithm}
\usepackage{eucal}
\usepackage{amsmath}
\usepackage{amssymb}
\usepackage[hang,flushmargin]{footmisc} 

\newcommand\blfootnote[1]{%
  \begingroup
  \renewcommand\thefootnote{}\footnote{#1}%
  \addtocounter{footnote}{-1}%
  \endgroup
}

\makeatletter
\newcommand*\bigcdot{\mathpalette\bigcdot@{.9}}
\newcommand*\bigcdot@[2]{\mathbin{\vcenter{\hbox{\scalebox{#2}{$\m@th#1\bullet$}}}}}
\makeatother

\usepackage{etoolbox}
\patchcmd{\quote}{\rightmargin}{\leftmargin 0.8em \rightmargin}{}{}

\usepackage[T1]{fontenc}
\usepackage[utf8]{inputenc}

\usepackage{microtype}

\title{Curriculum Script Distillation for Multilingual Visual Question Answering}

\author{Khyathi Raghavi Chandu \textsuperscript{\rm \textbf{1}}\\
  Allen Institute of AI \\
  \texttt{khyathic@allenai.org} \\\And
  Alborz Geramifard \\
  Meta \\
  \texttt{alborzg@meta.com} \\}

\begin{document}
\maketitle
\blfootnote{Work done while working at Meta}

\begin{abstract}

Pre-trained models with dual and cross encoders have shown remarkable success in propelling the landscape of several tasks in vision and language in Visual Question Answering (VQA). However, since they are limited by the requirements of gold annotated data, most of these advancements do not see the light of day in other languages beyond English. We aim to address this problem by introducing a curriculum based on the source and target language translations to finetune the pre-trained models for the downstream task. Experimental results demonstrate that script plays a vital role in the performance of these models. Specifically, we show that target languages that share the same script perform better ($\sim$ 6\%) than other languages and mixed-script code-switched languages perform better than their counterparts ($\sim$ 5-12\%).

\end{abstract}

\section{Introduction}

The remarkable ability of pre-trained language models has demonstrated their competence in several downstream tasks \cite{DBLP:conf/naacl/DevlinCLT19, DBLP:conf/interspeech/SchneiderBCA19}. The same trend continued in vision-and-language tasks including Visual Question Answering (VQA) \cite{DBLP:conf/emnlp/TanB19, DBLP:conf/nips/LuBPL19, zellers2021merlot}. While the benefits of the tasks are indisputable ranging from assisting visually impaired people to situated interactions, these advancements are limited to a single language. On top of this, these models are also not inherently capable of predicting code-mixed questions.  

However, these models are more than ever data hungry and demand large-scale multilingual vision and language datasets to achieve similar performance as demonstrated in English. The rapid advancements in monolingual vision-and-language tasks along with the strides in translation models can be leveraged to combine their powers for cross-lingual VQA \cite{DBLP:journals/corr/abs-2202-07630}. This brings the benefit of not needing to create parallel annotated  data which is expensive, as long as there is a translation model that can create weak translations to this target language.

To address this cross-lingual transfer, we perform data augmentation based on self-training to fine-tune multimodal models for the given task in a target language. Specifically, this repetition of fine-tuning is designed as a curriculum from the source language, to a concatenation of source and target languages, to finally predicting in the target language. The intermediate step of concatenating with the target language is based on a loose translation from a pre-trained unimodal translation model.

The preliminary results with different cross-modal encoders and dual encoders show that the performance of the target languages that share the same script as that of the pre-training languages is higher compared to the languages with different scripts. Similarly, the performance on code-mixed languages with mixed script that is Romanized performs better than their monolingual counterparts.

The main contributions of this paper are:

\begin{enumerate}
    \item We introduce a curriculum learning strategy for modeling multilingual questions with access to gold-annotated data in a single language. The curriculum is based on scheduling different languages between the source and the target in multiple stages of finetuning. We devise experiments to compare the performance of dual and cross encoders in monolingual and code-switched scenarios.
    \item We show that the models fare relatively well in the target languages that share the same script as that of English, i.e., Roman script in comparison to languages with different scripts. 
    \item We experimentally demonstrate that the partially Romanized context in code-switched cases contributes to increased performance compared to their monolingual counterparts. 
\end{enumerate}

This study is preliminary work on exploring the curriculum based on scripts of the source and the target languages. Based on these results, we believe that the curriculum of scripts has a lot more potential to explore.

\section{Curriculum Script Distillation}

\begin{figure}[t!]
\centering
\includegraphics[trim=3cm 4cm 9cm 8cm,clip,width=0.99\linewidth]{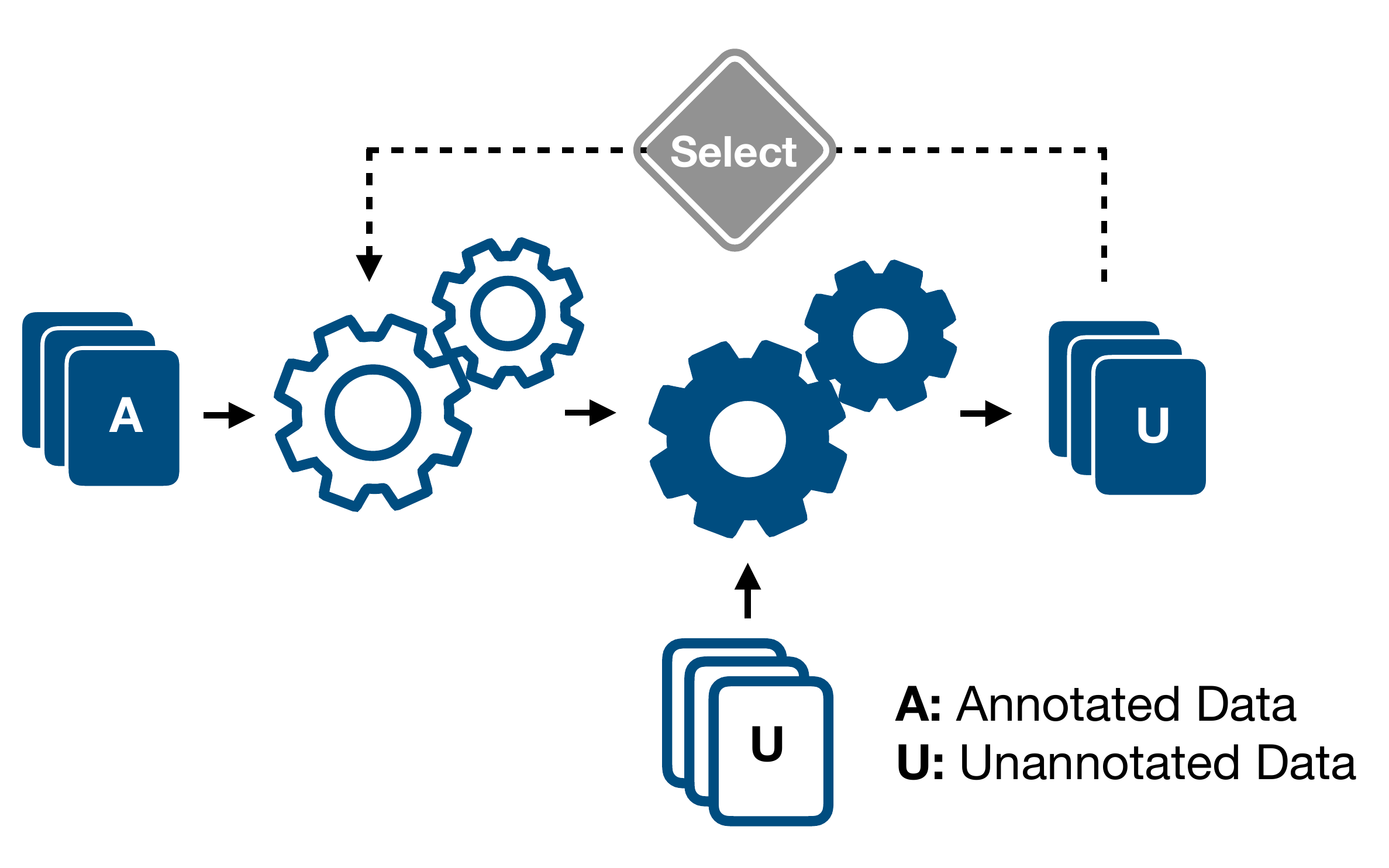}
\caption{Overall training setup}%\textcolor{red}{bigger fonts :)}}
\label{fig:model}
\end{figure}

The main idea of this method is to distill task-specific external knowledge from a large pre-trained English model to improve the performance of VQA in non-English languages. Specifically, we introduce a curriculum based on the combination of the scripts and languages of the source and target languages. The overall approach is presented in Figure \ref{fig:model}. We experiment with multiple pre-trained models and the core pre-trained model is abstracted out in this Figure. The algorithm is presented clearly in Algorithm \ref{alg:fst}. We select a pre-trained model $\CMcal{A}(\theta$), parameterized by $\theta$, and annotated data in any one language, which is English in our case. 

\begin{algorithm*}
\small
%\SetAlgoLined
\textbf{Input: } Pre-trained Model $\CMcal{A}(\theta$), Labeled Data in English 
$\CMcal{L}^{e}$, Translation Model $\CMcal{T}(\alpha$), Unlabeled Data $\CMcal{U}^{x}$ \\
\textbf{Output: } Trained End-Task Model $\CMcal{A}(\phi$) (finetuned on $\CMcal{A}$) \\
1. $\CMcal{A}(\theta^{'}$) $\leftarrow$ Fine-tune model $\CMcal{A}(\theta$) on $\CMcal{L}^{e}$ \\
2. $\CMcal{W}^{t}$ $\leftarrow$ Translate $\CMcal{L}^{e}$ to the target language using $\CMcal{T}(\alpha$) \\
3. $\CMcal{L}^{t}$ $\leftarrow$ Label answers to $\CMcal{W}^{t}$ using $\CMcal{A}(\theta^{'}$) on concatenated $\CMcal{L}^{e}$ $\bigoplus$ $\CMcal{W}^{t}$ \\
4. $\mathbb{L}^{t}$ $\leftarrow$ Subselect answers with high confidence from $\CMcal{L}^{t}$ \\
5. $\CMcal{A}(\phi$) $\leftarrow$ Fine-tune model $\CMcal{A}(\theta$) with $\mathbb{L}^{t}$ 
%6. Repeat Steps 3 to 5 by updating A($\theta^{''}$)  \\
 \caption{\small Self Training for Cross-lingual VQA }
 \label{alg:fst}
\end{algorithm*}

\paragraph{Task Finetuning: } The first step is to finetune a selected pre-trained model with annotated data in one language where the task-specific annotations are available. Owing to the ease of accessibility of English-paired annotations, this task-specific finetuning is performed on English data. This step adapts the model to the task in one language, which is not necessarily the target language that we want to test the performance on. We now have a pre-trained language model finetuned for a specific task. The initial pre-trained model $\CMcal{A}(\theta$) is now parameterized by the VQA specific parameters $\theta^{'}$ resulting in the model $\CMcal{A}(\theta^{'}$). Task finetuning is at one end of the spectrum where the languages or scripts from the target language remain unused, as it is completely based on the source language.

\paragraph{Cross-lingual Weak Annotation:} 
The next step is to adapt the model to the target language, which requires annotations from the target language. In order to annotate them, we first need questions in the target language. We  automatically translate them into the target language. The labeled examples in English $\CMcal{L}^{e}$ are translated to the target language, resulting in $\CMcal{W}^{t}$ which are weakly labeled translations in the target language. Note that the translations done here are unimodal machine translations rather than multimodally translated questions. These target language questions are concatenated with their English counterparts to assist the model adjust its parameters to the target language. The fine-tuned model from the previous step is used to annotate the concatenated questions along with an image to arrive at answer predictions. The English questions primed to the model act as a prompt to answer the queries by concatenating $\CMcal{L}^{e}$ with the distantly translated $\CMcal{W}^{t}$ questions. Then we use the previously fine-tuned A($\theta^{'}$) to predict the answers to the questions $\CMcal{W}^{t}$ to form $\CMcal{L}^{t}$. This falls in the middle ground of the two ends of the translation spectrum. The first end is using the fine-tuned model of English to predict the answers to questions from another language. The second end is using an off-the-shelf tool to translate the questions into the target language, which is used to fine-tune the model. While the first carries the language-uninformed zero-shot prediction, the second approach carries the burden of translation biases. We balance both of these cons in our approach by including the target language to enrich target language prediction beyond zero-shot and by priming with the question from the source language to counter translationese artifacts.

\paragraph{Selection: } We now have weakly translated questions and weakly annotated answers for the subset of the above questions. Instead of using the entire data, we select $\sim$ 5k questions which are high confidence answers i.e, among the weakly predicted answers above $\CMcal{L}^{t}$, we select $\mathbb{L}^{t}$ high confidence predictions. This weakly annotated data is added to the training data and model A($\theta^{'}$) is fine-tuned again to arrive at the final model $\CMcal{A}(\phi$). In our continued experiments, we repeated this entire process multiple times; however, we observed the results after the first iteration.

\section{Experiments and Results}

\paragraph{Dataset: } We conduct experiments on MuCO VQA dataset \cite{DBLP:conf/emnlp/KhanGE21} (Multilingual and Code-mixed) which is built on top of the VQA dataset \cite{antol2015vqa}. As the name suggests, the data includes both multilingual code-mixed questions and the answers remain in English. While our approach can be extended to a generation-based model, we plan to explore that in the future. With the discriminative-based model, the answers are present in English. Including English, the dataset includes questions from 6 different mono-languages and 5 different code-mixed languages. All the mixed languages are mixed with English. Aside from English, the other languages in the dataset are German, Spanish, French, Hindi, and Bengali. The monolingual translations are automatically obtained from Google Machine Translation, and the code-mixed questions are generated using the matrix language frame theory \cite{myers1997duelling}. 

\paragraph{Experiments: }
We combined experiments to analyze performance using multiple cross and dual encoders.

\noindent $\bigcdot$ \textbf{Cross-encoders:} The cross-encoder models use the image regions and language features to combine with a self-attention capable of discovering the alignments between the visual and the textual descriptions implicitly. Oftentimes, these models are trained with variants of masked language modeling extended to visual representations such as masked region classification.

\noindent $\bigcdot$ \textbf{Dual-encoders: } In contrast to cross-encoding where the inputs from both the modalities or languages are processed by the same encoder, dual encoders have 2 distinct encoders with separate parameters to process different modalities. The latent representations are then fused with late fusion techniques or extended to task specific model architectures. 

For pre-trained cross-encoders, we experimented with ViLT \cite{DBLP:conf/icml/KimSK21}, VisualBERT \cite{DBLP:journals/corr/abs-1908-03557}, LXMERT \cite{DBLP:conf/emnlp/TanB19}, CLIPViT. For pre-trained dual-encoders, we performed a sweep on combinations of visual encoders including ViT \cite{DBLP:conf/iclr/DosovitskiyB0WZ21}, BEiT \cite{DBLP:conf/iclr/Bao0PW22}, DEiT \cite{DBLP:conf/icml/TouvronCDMSJ21} 
and textual encoders including XLM \cite{DBLP:conf/acl/ConneauKGCWGGOZ20}, XGLM \cite{DBLP:journals/corr/abs-2112-10668}, mT5 \cite{DBLP:conf/naacl/XueCRKASBR21}, and CANINE \cite{DBLP:journals/tacl/ClarkGTW22}.

We used the official evaluation script provided by the VQA challenge. 

%textual encoders like xlm, xglm, mt5, xroberta, canine with a ViT based feature extractor. 

\begin{figure}[t!]
\centering
\includegraphics[trim=8cm 5cm 10cm 10cm,clip,width=0.99\linewidth]{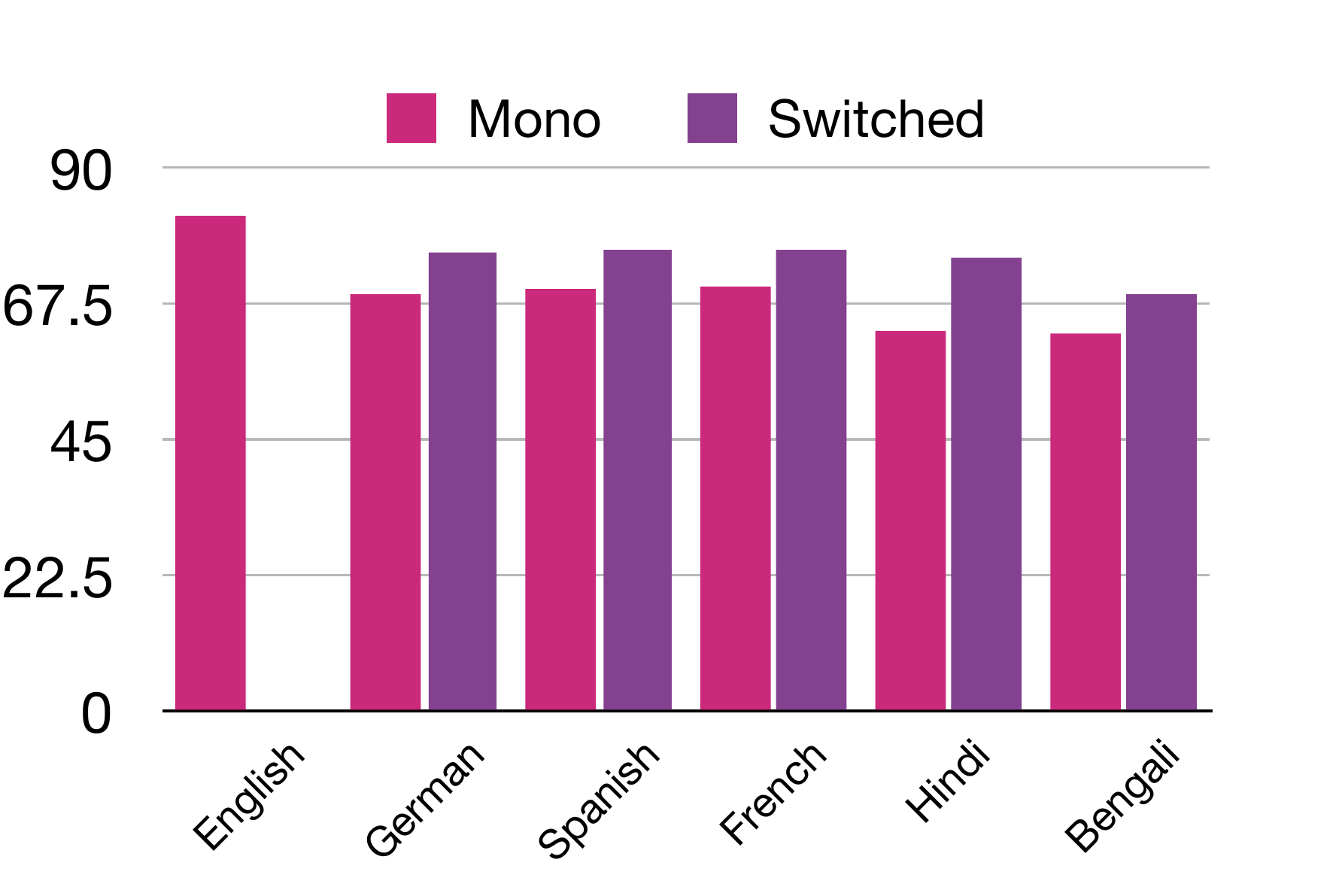}
\caption{Overall accuracy of VQA across multiple monolingual and code-switched questions}
\label{fig:results}
\end{figure}

\paragraph{Results: }
The trends of the results across languages are shown in Figure \ref{fig:results}. We ran each model 5 times for each language and the scores were averaged across the models and the runs.

From our preliminary results with cross and dual encoders, performance with Roman scripts  is better than those with non-roman scripted languages. Across the aforementioned model configurations, the performance between Roman and non-Roman languages has a  minimum difference of at least $\sim$ 6\%. Since the base model is primarily pre-trained on English which is in the Roman script, this preferred trend continues in other languages as well. Naturally, an open question arises if the performance gap can be reduced with increasing scripts from other languages during pre-training, which does not mandate expensive parallel data.

Additionally, the performance on switched concepts in English with code-switching is better than their counterparts, (by about $\sim$ 5-12\%). This shows that the  knowledge from large English pretrained language model can be constructively used to improve performance in other languages. Our model selects the data based on the performance of the first fine-tuned model on the combined questions of the source and the target languages, thereby selecting examples that are performing well on mixed yet combined questions. The natural extension here is to understand the performance based on the selection of instances performing better in one language but not the other, which we plan to explore in the future.

\section{Related Work}

\paragraph{Multilingual VQA: } Several research works extended existing monolingual VQA dataset to other languages \cite{raj-khan-etal-2021-towards-developing, DBLP:conf/ijcnlp/GuptaLEB20}. Similar to these efforts, \citet{ramnath-etal-2021-worldly} also generated synthetic data to perform fact-based spoken VQA. While the above datasets are synthetically generated, \citet{DBLP:conf/nips/GaoMZHWX15, shimizu-etal-2018-visual} crowdsourced datasets directly in languages other than English.

\paragraph{Multilingual Multimodal Modeling: } A detailed account of various modeling approaches was outlined categorically in the recent survey paper \cite{DBLP:journals/corr/abs-2210-16960}. They described categories of modeling approaches as modular, projecting embeddings, and mapping which includes popular approaches like alignment and attention. Our work mostly falls in the category of embedding projection. We build on the fusion-based approach as embedding projection of the cross-attention to schedule a curriculum of language mixing, by concatenating the source language with the target language. This approach is inspired by the translation objective introduced by \citet{DBLP:conf/nips/ConneauL19}. The knowledge from one language is distilled to train a student network to perform visual question answering by \citet{DBLP:conf/emnlp/KhanGE21}. The student network that learns to perform the task in several languages is trained to imitate the teacher network that is trained on English-annotated data at all the intermediate layers instead of the final layer alone.

\paragraph{Augmentation with self-training: } 
Prior research widely used annotation of data with human or weak distant labels to use as augmented data for training in low-resource settings. In some cases, the additional data is augmented not in the form of end task but to provide additional information about the context such as providing object labels \cite{DBLP:conf/coling/ChanduSCTS22, gupta-etal-2021-vita}.
\citet{DBLP:conf/emnlp/ChopraRBC21} performed self-training with switch-point based selection of instances to repurpose large pre-trained models to code-switched sequence labeling tasks. Instead of selecting biased examples based on the poor-performing switch point, we select the examples that the model performs well on  the concatenated sentences of the source and the target languages. 

\section{Conclusions and Future Work}
Relaying the technologies developed in high-resource languages to several languages other than English has always taken a back seat. A critical reason hindering the smooth relay is the lack of annotated resources of the same scale as that of English in other languages. This is particularly evident in resource-heavy tasks that include multimodal tasks such as Visual Question Answering. 
However, independent of these multimodal tasks, machine translation has made commendable strides in the field. Our work leverages these unimodal machine translation models to build a curriculum of the scripts and languages in the training process. Specifically, we performed self-training on a pre-trained multimodal model to learn the task-specific parameters first on the source language, then use a concatenated source and weakly translated target language question to make the predictions. Using these answer predictions as weak labels, the model was finetuned for the target language and eventually tested on the target language. In this way, we are not relying on the ground truth gold annotations of the target language, but rather seek help from the source language of the questions to predict answers for the target language questions. Our experimental results emphasize two main conclusions. First, the performance on languages with a similar script as that of the languages in the pretrained model is better than different scripts. Second, the performance on code-switched languages is better than their counterparts. Our work presents preliminary results on the potential of mixed scripts in training multilingual and cross-lingual models. We plan to explore fine-grained improvements contributed by scripts on dual and cross encoders in different stages of self-training in the future.

\bibliography{anthology,custom}
\bibliographystyle{acl_natbib}

\pagebreak
\newpage
\clearpage 

\appendix

% \section{Example Appendix}
% \label{sec:appendix}

% This is an appendix.

\end{document}